\documentclass{article}

\usepackage{caption}
\usepackage{wrapfig}
\usepackage{listings}
\usepackage[most]{tcolorbox}

    \usepackage[preprint]{neurips_2025}

\usepackage[utf8]{inputenc} % allow utf-8 input
\usepackage[T1]{fontenc}    % use 8-bit T1 fonts
\usepackage{hyperref}       % hyperlinks
\usepackage{url}            % simple URL typesetting
\usepackage{booktabs}       % professional-quality tables
\usepackage{amsfonts}       % blackboard math symbols
\usepackage{nicefrac}       % compact symbols for 1/2, etc.
\usepackage{microtype}      % microtypography
\usepackage{xcolor}         % colors

\usepackage{graphicx}

\definecolor{jsonbackground}{RGB}{248, 248, 248}
\definecolor{jsonkey}{RGB}{28, 126, 214}
\definecolor{jsonstring}{RGB}{196, 26, 22}
\definecolor{jsonnumber}{RGB}{9, 134, 88}
\definecolor{jsonkeyword}{RGB}{170, 93, 0}

\lstdefinelanguage{json}{
    keywords={false,true,null},
    keywordstyle=\color{jsonkeyword}\bfseries,
    sensitive=false,
    comment=[l]{//},
    morecomment=[s]{/*}{*/},
    commentstyle=\color{gray}\ttfamily,
    stringstyle=\color{jsonstring},
    morestring=[b]",
    morestring=[b]',
}

\lstdefinestyle{json}{
    language=json,
    basicstyle=\ttfamily\small,
    numbers=none,              % 不显示行号
    showstringspaces=false,
    breaklines=true,
    frame=single,              % 单线边框
    frameround=tttt,           % 圆角边框
    framexleftmargin=0pt,      % 减少左边框内边距
    framexrightmargin=8pt,
    framextopmargin=4pt,
    framexbottommargin=4pt,
    backgroundcolor=\color{jsonbackground},
    rulecolor=\color{gray!30},
    xleftmargin=0em,         % 减少整体左边距
    xrightmargin=1em,
    aboveskip=1em,
    belowskip=1em,
    literate=
        *{0}{{{\color{jsonnumber}0}}}{1}
        {1}{{{\color{jsonnumber}1}}}{1}
        {2}{{{\color{jsonnumber}2}}}{1}
        {3}{{{\color{jsonnumber}3}}}{1}
        {4}{{{\color{jsonnumber}4}}}{1}
        {5}{{{\color{jsonnumber}5}}}{1}
        {6}{{{\color{jsonnumber}6}}}{1}
        {7}{{{\color{jsonnumber}7}}}{1}
        {8}{{{\color{jsonnumber}8}}}{1}
        {9}{{{\color{jsonnumber}9}}}{1}
        {:}{{{\color{black}:}}}{1}
        {,}{{{\color{black},}}}{1}
        {\{}{{{\color{black}\{}}}{1}
        {\}}{{{\color{black}\}}}}{1}
        {[}{{{\color{black}[}}}{1}
        {]}{{{\color{black}]}}}{1},
}

\title{VIS-Shepherd: Constructing Critic for LLM-based Data Visualization Generation}

\author{%
  Bo Pan \\
  State Key Lab of CAD\&CG\\
  Zhejiang University\\
  \texttt{bopan@zju.edu.cn} \\
  \And
  Yixiao Fu \\
  State Key Lab of CAD\&CG\\
  Zhejiang University\\
\texttt{3210101100@zju.edu.cn} \\
  \AND
  Ke Wang \\
  State Key Lab of CAD\&CG\\
  Zhejiang University\\
\texttt{sttotphd@zju.edu.cn}
  \And
  Junyu Lu \\
  State Key Lab of CAD\&CG\\
  Zhejiang University\\
\texttt{lujunyu@zju.edu.cn}
  \And
  Lunke Pan \\
  State Key Lab of CAD\&CG\\
  Zhejiang University\\
\texttt{3230102935@zju.edu.cn}
  \And
  Ziyang Qian \\
  State Key Lab of CAD\&CG\\
  Zhejiang University\\
\texttt{qian\_ziyang@zju.edu.cn}
  \And
  Yuhan Chen \\
  State Key Lab of CAD\&CG\\
  Zhejiang University\\
\texttt{3230105086@zju.edu.cn}
  \And
  Guoliang Wang \\
  State Key Lab of CAD\&CG\\
  Zhejiang University\\
\texttt{wangglr@zju.edu.cn}
  \And
  Yitao Zhou \\
  State Key Lab of CAD\&CG\\
  Zhejiang University\\
\texttt{3220105480@zju.edu.cn}
  \And
  Li Zheng \\
  State Key Lab of CAD\&CG\\
  Zhejiang University\\
\texttt{zhengli@zju.edu.cn}
  \And
  Yinghao Tang \\
  State Key Lab of CAD\&CG\\
  Zhejiang University\\
\texttt{yinghaotang@zju.edu.cn}
  \And
  Zhen Wen \\
  State Key Lab of CAD\&CG\\
  Zhejiang University\\
\texttt{wenzhen@zju.edu.cn}
  \And
  Yuchen Wu \\
  State Key Lab of CAD\&CG\\
  Zhejiang University\\
\texttt{3230100789@zju.edu.cn}
  \And
  Junhua Lu \\
  Hangzhou Research Institute \\ of AI and Holographic Technology \\
\texttt{akiori@zju.edu.cn}
  \And
  Biao Zhu \\
  Hangzhou Research Institute \\ of AI and Holographic Technology \\
\texttt{zhubiao@zju.edu.cn}
  \And
  Minfeng Zhu \\
  Zhejiang University\\
\texttt{minfeng\_zhu@zju.edu.cn}
  \And
  Bo Zhang \\
  State Key Lab of CAD\&CG\\
  Zhejiang University\\
  \texttt{bo.zhang@zju.edu.cn} \\
  \And
  Wei Chen \\
  State Key Lab of CAD\&CG\\
  Zhejiang University\\
  \texttt{chenvis@zju.edu.cn} \\
}

\begin{document}

\maketitle

\begin{abstract}
Data visualization generation using Large Language Models (LLMs) has shown promising results but often produces suboptimal visualizations that require human intervention for improvement. In this work, we introduce VIS-Shepherd, a specialized Multimodal Large Language Model (MLLM)-based critic to evaluate and provide feedback for LLM-generated data visualizations. At the core of our approach is a framework to construct a high-quality visualization critique dataset, where we collect human-created visualization instances, synthesize corresponding LLM-generated instances, and construct high-quality critiques. 
We conduct both model-based automatic evaluation and human preference studies to evaluate the effectiveness of our approach. Our experiments show that even small (7B parameters) open-source MLLM models achieve substantial performance gains by leveraging our high-quality visualization critique dataset, reaching levels comparable to much larger open-source or even proprietary models. Our work demonstrates significant potential for MLLM-based automated visualization critique and indicates promising directions for enhancing LLM-based data visualization generation. Our project page: \url{ https://github.com/bopan3/VIS-Shepherd}.
\end{abstract}

\section{Introduction}
Data visualization plays an essential role in data analysis and communication across various domains \cite{Semiology_of_graphics, wilkinson2011grammar, munzner2014visualization}. Large Language Models (LLMs) have recently demonstrated remarkable potential for automating visualization generation by interpreting natural language instructions and producing corresponding visualization code \cite{matplotagent, lida, chartllama}. However, despite these advances, state-of-the-art LLMs still frequently generate deficient data visualizations \cite{li2024visualizationgenerationlargelanguage, VisEval}. 
These shortcomings highlight the need for automatic evaluation and feedback mechanisms to ensure high-quality visualizations.

Recent works have explored both general-purpose critics that provide broad feedback \cite{wang2023shepherdcriticlanguagemodel, UltraFeedbackBL, critiquellm} and domain-specific critics that offer specialized evaluation in fields such as mathematics \cite{Math-Shepherd}, programming \cite{Weyssow2024CodeUltraFeedbackAL}, and medicine \cite{gupta2025medcodemedicalcritiquebased}. However, existing methods fall short when applied to data visualization generation, which presents unique challenges. First, critics that solely rely on textual input are insufficient for visualization evaluation.
Visualization generation differs fundamentally from tasks like text or code generation in that the final output is a rendered visual representation. Second, visualization assessment requires specialized domain knowledge—even humans require training and practical experience to effectively identify defects in visualization results. These challenges highlight the need for visualization-specific critics that can systematically identify and evaluate visualization-specific defects based on established principles and empirical understanding of visualization quality. 

\begin{figure*}[t]
        \centering
        \includegraphics[width=1.0\textwidth]{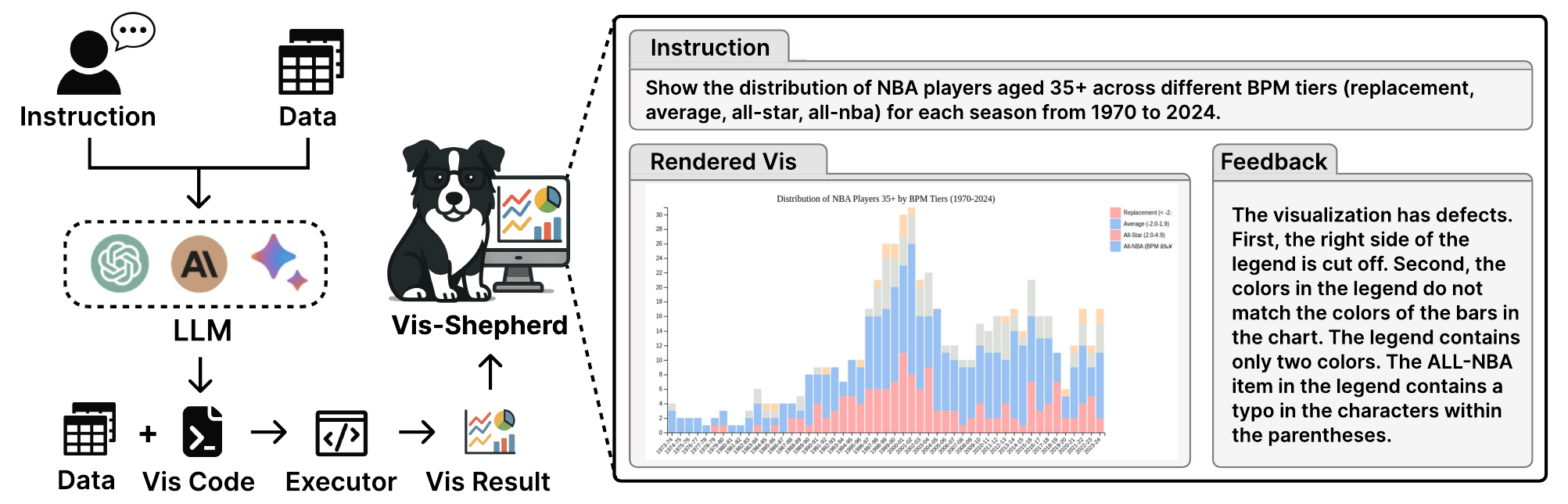}
    \caption{Overview of the VIS-Shepherd framework. The diagram illustrates the end-to-end visualization generation and critique process. Starting with user-provided instructions and data, an LLM generates corresponding visualization code. This code is executed to produce a visualization result. VIS-Shepherd then critiques the output visualization, identifying defects and providing feedback.}
    \label{fig:overview}
\end{figure*}

In this work, we present VIS-Shepherd, a Multimodal Large Language Model (MLLM)-based critic specialized for evaluating and providing feedback for data visualization generation. Figure \ref{fig:overview} illustrates how VIS-Shepherd works in the visualization generation pipeline. Given a dataset and a user-provided instruction, an LLM generates the corresponding visualization code, which is then executed to obtain the visualization result. VIS-Shepherd then critically assesses the output, providing constructive feedback by identifying specific defects and offering suggestions for improvement.

To develop VIS-Shepherd, we propose a systematic framework for constructing a high-quality visualization critique dataset.
Our framework consists of a multi-stage process: First, to collect diverse real-world visualization scenarios and high-quality visualization instances, we collected 180K visualization instances from the internet and carefully curated a high-quality subset of 1.7K visualization instances. Second, to simulate real-world generation scenarios, we designed a customized LLM-based pipeline that synthesizes user instructions and exports corresponding datasets for these visualizations, resulting in (instruction, data, human-created high-quality visualization) triplets. Third, to synthesize a dataset showcasing various defects in LLM-generated visualizations, we employed state-of-the-art LLMs to generate visualizations based on the same instructions, extending our dataset to (instruction, data, human-created high-quality visualization, LLM-generated visualization) quadruples. Finally, we recruited annotators with visualization expertise to carefully identify defects and provide critiques of the LLM-generated visualizations by comparing them against their human-created counterparts, resulting in 2.7k high-quality visualization critiques. This carefully curated dataset enables VIS-Shepherd to learn from real-world visualization expertise and provide constructive feedback on LLM-generated visualizations.

To verify the effectiveness of our framework, we fine-tuned a 7B-parameter open-source MLLM on our dataset and conducted both model-based automated evaluation and human preference studies. The evaluation results demonstrate that our fine-tuned model achieves substantial improvements in visualization critique quality. Notably, despite its relatively compact size, our model attains comparable performance to significantly larger (e.g., $10\times$) state-of-the-art open-source and proprietary models, highlighting the effectiveness of our approach. 

In summary, this work makes the following contributions: (1) We propose VIS-Shepherd, a specialized MLLM-based critic to evaluate and provide feedback on LLM-generated data visualizations; (2) We propose a systematic framework for constructing high-quality visualization critique datasets and contribute a dataset of 2.7k high-quality visualization critiques; (3) We conduct both automatic and human evaluations demonstrating that our specialized 7B-parameter model achieves performance comparable to significantly larger state-of-the-art models, validating the effectiveness of our approach.

\section{Related Work}
In this section, we review the existing literature on LLM-based visualization generation and critique models, contextualizing our work within these complementary research domains.
\subsection{LLM-based Data Visualization Generation}

Benefiting from their powerful natural language understanding and code generation capabilities, Large Language Models (LLMs) have demonstrated significant potential in data visualization generation. Existing approaches have enhanced visualization quality through prompt engineering techniques \cite{chat2vis, lida, multimodalprompt}, fine-tuning methods \cite{chartllama, chartgpt}, and agent-based frameworks \cite{matplotagent, vispath, nvagent}. Prompt engineering has proven effective in systems like Chat2VIS \cite{chat2vis}, which transforms natural language queries directly into visualization code, and LIDA \cite{lida}, which employs a sophisticated multi-stage pipeline for data summarization and visualization goal exploration. To address domain-specific needs, fine-tuning approaches such as ChartLlama \cite{chartllama} have emerged, training multimodal LLMs on custom datasets for improved chart understanding, while ChartGPT \cite{chartgpt} breaks down the generation process into step-by-step reasoning to handle abstract natural language inputs. Taking complexity management further, agent-based frameworks like MatPlotAgent \cite{matplotagent} integrate query understanding and iterative debugging modules, and VisPath \cite{vispath} employs multi-path reasoning with feedback optimization to tackle underspecified queries. Our work takes a different yet equally important direction by focusing on developing a specialized critic model based on multimodal LLMs that evaluates visualizations and provides actionable feedback, offering an orthogonal contribution that can complement and enhance all these existing approaches.

\subsection{Critique Models}
Critique models have emerged as an effective paradigm for enhancing LLMs by providing systematic evaluation and actionable feedback. General-purpose critics \cite{wang2023shepherdcriticlanguagemodel, UltraFeedbackBL, critiquellm, madaan2023self, liu2023g} evaluate content across multiple dimensions, demonstrating the efficacy of LLMs in assessing outputs regardless of domain or content type. More specialized domain-specific critics have been developed for fields including mathematics \cite{Math-Shepherd, golovneva2022roscoe}, programming \cite{sean2023generating,Weyssow2024CodeUltraFeedbackAL}, and healthcare \cite{gupta2025medcodemedicalcritiquebased}, where particular knowledge and constraints significantly influence quality assessment. Recent work has established evaluation methodologies for critique models, with CriticBench \cite{lin2024criticbench} providing a comprehensive benchmark that assesses LLMs' abilities to critique and rectify reasoning across both general and specialized applications. Despite these advances, existing methods fall short when applied to data visualization generation. Our work addresses this gap by building MLLM-based critique models that systematically identify and evaluate visualization defects based on established principles and empirical understanding of visualization quality.
\section{Dataset Construction}
\definecolor{mygray}{RGB}{140,140,140} % 你可以调整RGB值

\begin{table*}[t]
% \vspace{-10pt}
\begin{center}
\renewcommand{\arraystretch}{1.2}
\fontsize{11pt}{13pt}\selectfont
\scalebox{0.57}
{
\begin{tabular}{@{}p{114pt}p{320pt}p{103pt}p{103pt}@{}}
\toprule
% \cellcolor{white}
{\textbf{Defect Type}}
& {\textbf{Example Visualization}} \
& {\textbf{Example good critique}} \
& {\textbf{Example bad critique}} \ \
\\ \midrule

Instruction Compliance: 

The candidate visualization fails to meet all requirements specified in the user instructions.
& Instruction: Generate a multi-faceted line chart showing the 7-day average of COVID-19 cases for selected countries, highlighting the peak case numbers with markers and labels.  \par
% \textbf{Reference Visualization}: 
% \textbf{Candidate Visualization}: 
        \begin{minipage}{320pt}
            % 第一张图片
            \begin{minipage}
            {0.5\textwidth}
            \centering % 让标题和图片居中对齐
                \textbf{Reference Visualization} \\ % 第一张图片上方文字
                \includegraphics[width=\textwidth]{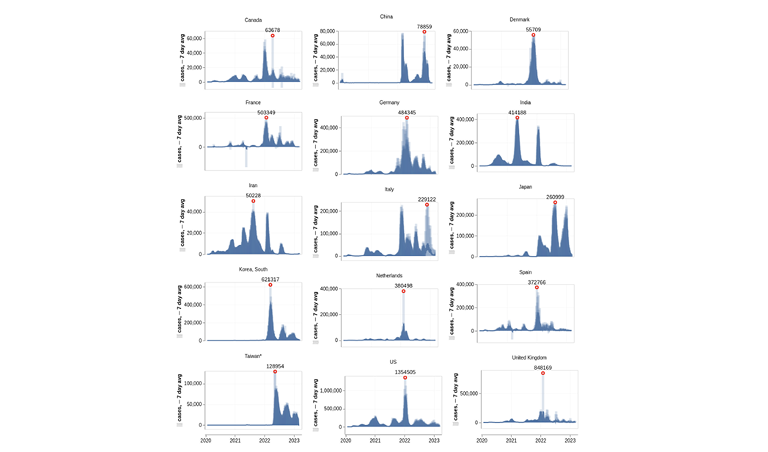} % 第一张图片
            \end{minipage}
            \hfill % 图片之间的间隔
            % 第二张图片
            \begin{minipage}{0.5\textwidth}
            \centering % 让标题和图片居中对齐
                \textbf{Candidate Visualization} \\ % 第二张图片上方文字
                \includegraphics[width=\textwidth]{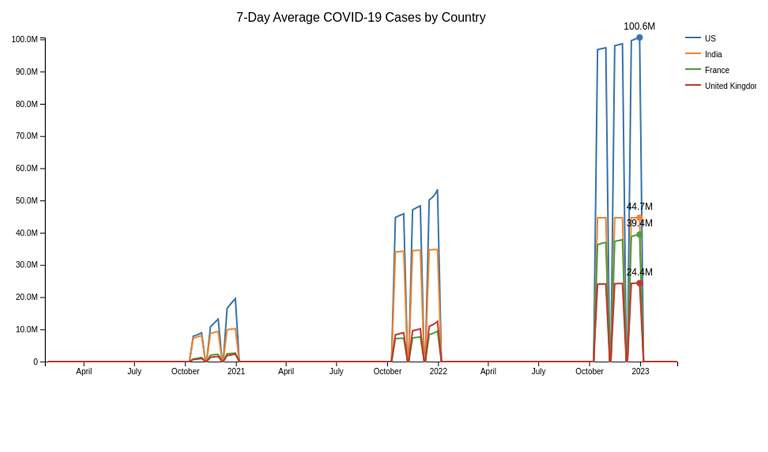} % 第二张图片
            \end{minipage}
        \end{minipage}
\par

& The "multi-faceted" requirement has not been met. Please split the data for each country into separate subplots to visualize them individually.  
& Separate subplots to visualize them individually. \par
\par \hspace{1cm} \par
\textcolor{mygray}{This critique is bad because there is only a suggestion but no highlighting of defects.}
\\\cline{1-4}

Visual Clarity:

The candidate visualization shows visual clarity issues (e.g., incomplete elements, undersized elements, overlapping, overcrowding).

& Instruction: Show me a visualization of pizza orders grouped by size and type, with the number of orders plotted over time.  \par
% \textbf{Reference Visualization}: 
% \textbf{Candidate Visualization}: 
        \begin{minipage}{320pt}
            % 第一张图片
            \begin{minipage}{0.5\textwidth}
            \centering % 让标题和图片居中对齐
                \textbf{Reference Visualization} \\ % 第一张图片上方文字
                \includegraphics[width=\textwidth]{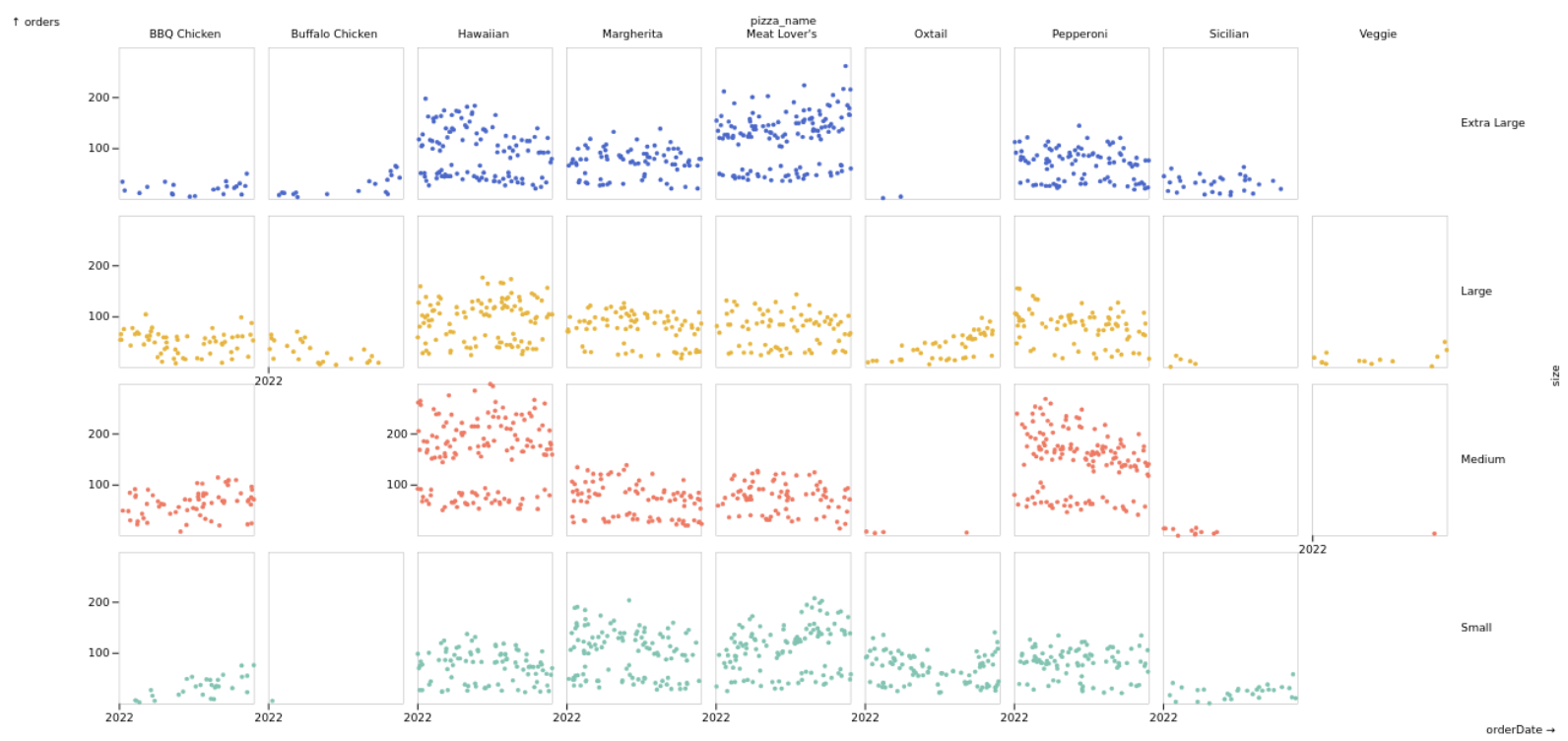} % 第一张图片
            \end{minipage}
            \hfill % 图片之间的间隔
            % 第二张图片
            \begin{minipage}{0.5\textwidth}
            \centering % 让标题和图片居中对齐
                \textbf{Candidate Visualization} \\ % 第二张图片上方文字
                \includegraphics[width=\textwidth]{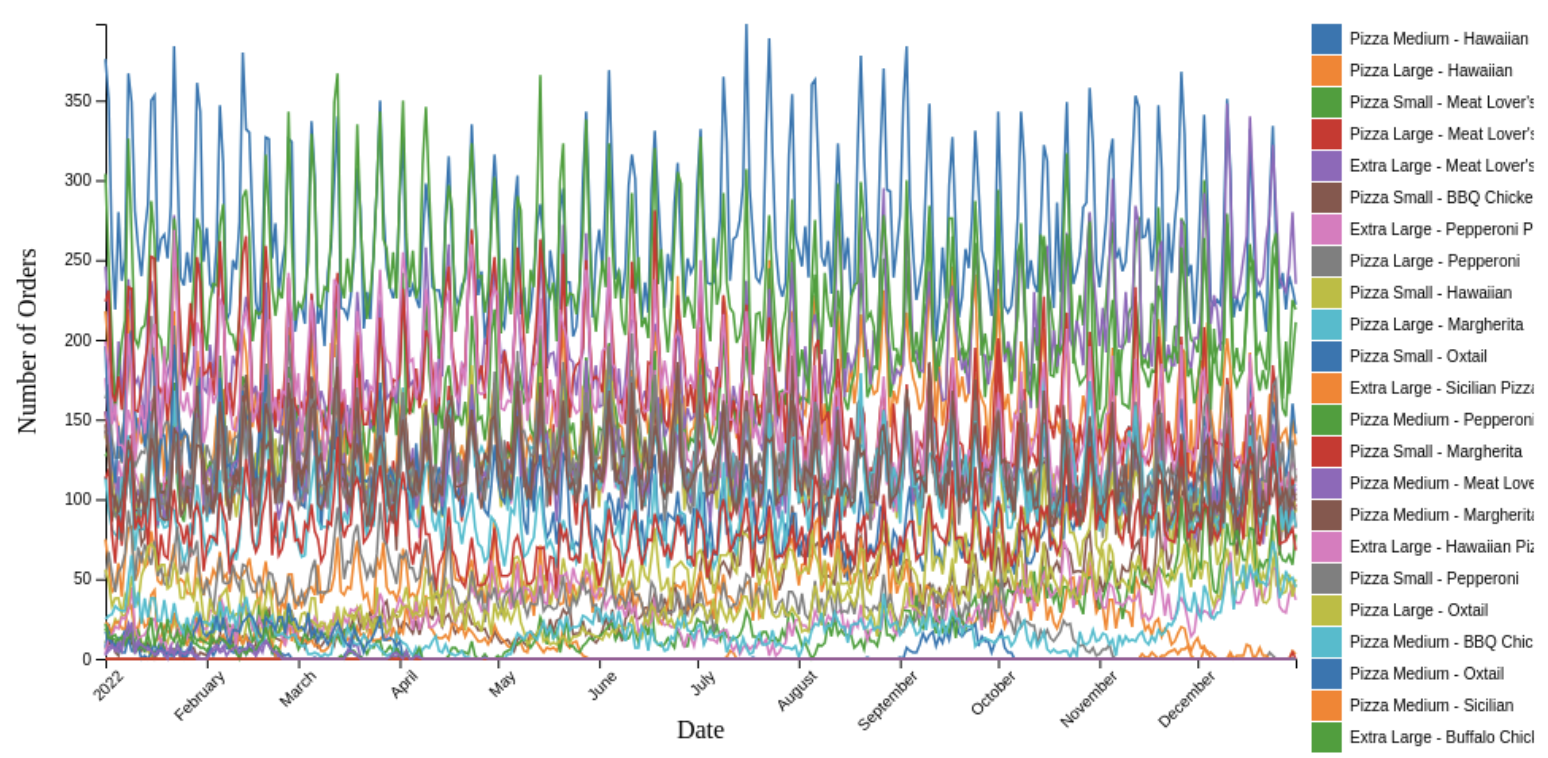} % 第二张图片
            \end{minipage}
        \end{minipage}
\par

& The line chart is too densely plotted, causing overlapping and obscured lines. Split the chart into separate subplots based on 'size' to achieve a clearer presentation. Moreover, the right side of the color legend is cutoff.  
& Bad visual clarity. \par
\par \hspace{1cm} \par
\textcolor{mygray}{This critique is bad because it just
points out the error type but doesn’t
mention where the error is.}
\\\cline{1-4}

Semantical Readability:

The candidate visualization contains visual elements with unclear meanings that cause confusion, and/or defective labeling that impairs the readability of element meanings.
& Instruction: Show the market share evolution of BMW electric vehicle models by model year and model, with the count of records represented by bubble size.  \par
% \textbf{Reference Visualization}: 
% \textbf{Candidate Visualization}: 
        \begin{minipage}{300pt}
            % 第一张图片
            \begin{minipage}{0.5\textwidth}
            \centering % 让标题和图片居中对齐
                \textbf{Reference Visualization} \\ % 第一张图片上方文字
                \includegraphics[width=\textwidth]{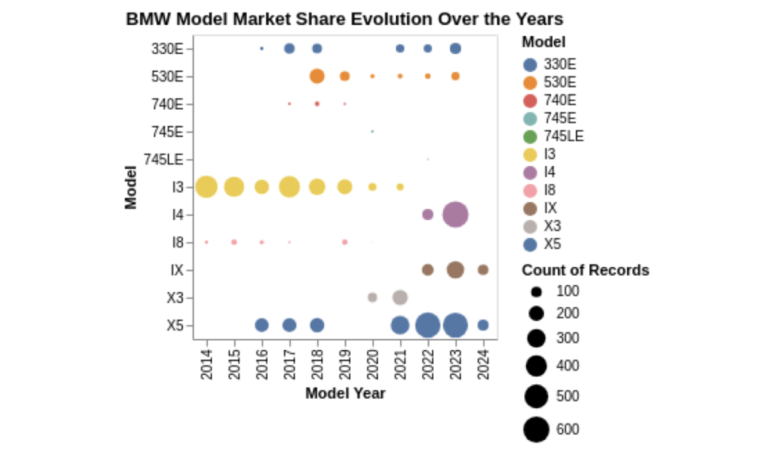} % 第一张图片
            \end{minipage}
            \hfill % 图片之间的间隔
            % 第二张图片
            \begin{minipage}{0.5\textwidth}
            \centering % 让标题和图片居中对齐
                \textbf{Candidate Visualization} \\ % 第二张图片上方文字
                \includegraphics[width=\textwidth]{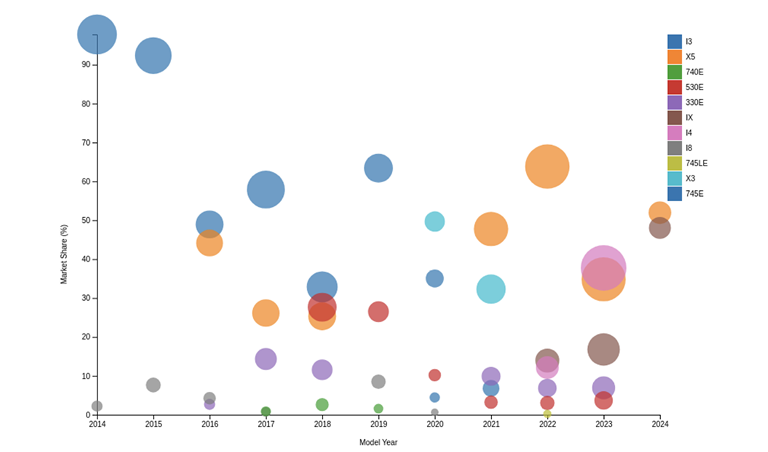} % 第二张图片
            \end{minipage}
        \end{minipage}
\par

& The meaning of bubble size is unclear. Add a size legend for it. The meaning of the items in the color legend is unclear. Add a legend title for it.

& Enlarge the size of texts to ease reading and use more professional fonts. \par
\par \hspace{1cm} \par
\textcolor{mygray}{This critique is bad because it just provide preference suggestions while ignores important semantic readability defects.}
\\\cline{1-4}

No Defect:

The candidate visualization has no defects for delivery. Provide preference suggestions to further improve the candidate visualization.
& Instruction: Create a treemap visualization showing the electricity generation by source for different countries, with each country represented as a parent node and sources as child nodes.  \par
% \textbf{Reference Visualization}: 
% \textbf{Candidate Visualization}: 
        \begin{minipage}{320pt}
            % 第一张图片
            \begin{minipage}{0.5\textwidth}
            \centering % 让标题和图片居中对齐
                \textbf{Reference Visualization} \\ % 第一张图片上方文字
                \includegraphics[width=\textwidth]{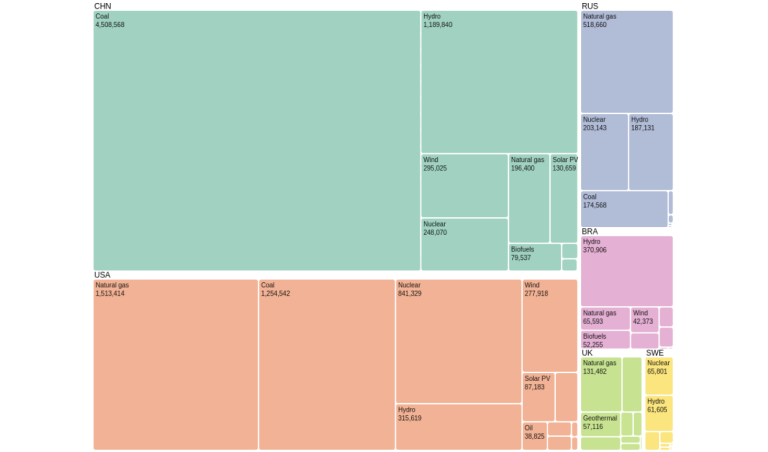} % 第一张图片
            \end{minipage}
            \hfill % 图片之间的间隔
            % 第二张图片
            \begin{minipage}{0.5\textwidth}
            \centering % 让标题和图片居中对齐
                \textbf{Candidate Visualization} \\ % 第二张图片上方文字
                \includegraphics[width=\textwidth]{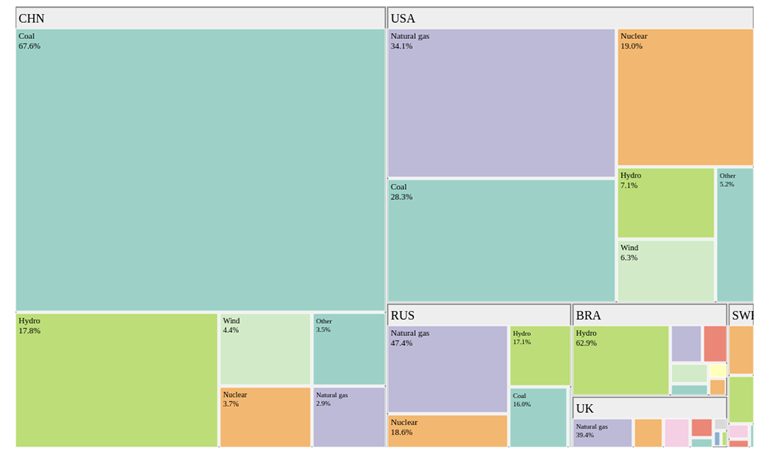} % 第二张图片
            \end{minipage}
        \end{minipage}
\par

& Current result has no defect. Suggestion: Improve the visual hierarchy by adding more prominent borders or spacing between different countries to better distinguish the country-level groupings.

& Current result has no defect. \par
\par \hspace{1cm} \par
\textcolor{mygray}{This critique is bad because it does not provide any suggestion for improvement when there is no defect.}
\\\cline{1-4}

\bottomrule
\end{tabular}
}
\end{center}
\caption{%\footnotesize 
Categories of common defects in LLM-generated visualizations with example critiques. This table presents a structured categorization of typical defects identified by visualization domain experts during preliminary analysis of 200 samples. For each defect type, the table provides example visualizations (both human-created reference and LLM-generated candidate), alongside demonstrations of effective critiques that identify specific issues and offer constructive solutions, contrasted with ineffective critiques that lack specificity or fail to properly address the actual defects. This categorization served as a guideline for annotators during the systematic critique collection process.
}
  \label{tab:human_examples}%
\end{table*}

The development of VIS-Shepherd required a high-quality dataset of visualization critiques, which was not previously available. To address this, we designed a systematic four-stage framework to construct such a dataset. This framework includes: (1) curation of human-created visualization instances; (2) instruction synthesis and dataset exportation; (3) LLM-based visualization generation; and (4) collection of high-quality critiques. This section details the technical aspects and methodologies of each stage.

\subsection{Stage 1: Human-created Instance Curation}
To effectively provide feedback on visualizations generated across diverse real-world scenarios, our dataset needed to reflect the variety of data types, visualization forms, and usage contexts found in practice. Existing NL2VIS datasets often focus on narrow domains \cite{text2vis, NLV} or rely on rule-based synthesis of limited visualization types \cite{VisEval, nvbench, nvbench2}, making them insufficient for our purposes.

To address this limitation, we crawled 180K real-world human-created visualization instances with code from Observable\footnote{https://observablehq.com/}, a public platform hosting a wide range of human-created visualization instances. Since these real-world instances vary significantly in quality, we implemented a multi-stage filtering process to ensure instances in our final dataset are of high-quality. First, we employed a closed-source MLLM \cite{gemini2023} to filter out visualizations that were of poor quality (e.g., blank outputs, severely incomplete representations, or those using toy datasets) or that contained interactive elements or animations, as our focus is on constructing a critic model for static visualizations. This initial filtering yielded 67K visualization instances.

Subsequently, we recruited human annotators with visualization expertise to manually inspect these visualizations and select those exhibiting high-quality, resulting in 6.9K high-quality instances. After conducting dataset export validation, code deduplication with simhash \cite{sadowski2007simhash}, and manual deduplication based on rendered outputs, we ultimately curated a final set of 1.7K diverse, high-quality, human-created visualization exemplars that serve as the foundation for our dataset construction process.

\subsection{Stage 2: Instruction Synthesis and Dataset Exportation}
In LLM-based visualization generation scenarios, the model typically requires both a user instruction expressing visualization intent and a preview of the dataset to be visualized \cite{VisEval}. For this stage, we designed a customized LLM-based pipeline to generate these elements from our curated in-the-wild visualization instances.

For instruction synthesis, we employed a role-playing approach \cite{ge2024scaling, lambert2025tulu3pushingfrontiers} where the LLM first imagined potential user backgrounds and usage scenarios for each visualization instance, then generated corresponding plausible user instructions. This approach ensured diversity in instruction styles, complexity levels, and domain-specific terminology, better reflecting real-world usage patterns.

For dataset exportation, we faced the challenge that in-the-wild visualization instances often store data using various methods and formats, sometimes even hard-coded directly within the visualization code. We addressed this by having the LLM generate customized dataset exportation scripts for each visualization instance. To validate data completeness and usability, we then had the LLM rewrite a version of the visualization code that achieved the same visual effect but read from the exported data instead of the original source.

Since all visualization data can theoretically be converted to a tidy data format \cite{wickham2014tidy}, we instructed the LLM to export data primarily in CSV format. For certain domain-specific data types with established standard formats (e.g., GeoJSON for geographic visualizations), we permitted exportation in appropriate corresponding formats. Following data exportation, we generated standardized preview information for each dataset using methods and formats similar to previous work \cite{lida, VisEval}, ensuring consistency in the information provided to visualization generation models.

This stage yielded structured triplets $(I, D, V_h)$ where $I$ represents the instruction, $D$ the dataset, and $V_h$ the human-created visualization, establishing the foundation for subsequent LLM-based visualization generation result collection.

\subsection{Stage 3: LLM-based Visualization Generation}
To capture the spectrum of visualization defects typically produced by state-of-the-art LLMs, we utilized advanced models like GPT-4o \cite{openai2024gpt4ocard}and Claude-Sonnet-3.5 \cite{anthropic2024claude} to generate visualizations ($V_{LLM}$) based on the instruction-dataset pairs $(I, D)$ from Stage 2. Given the inherent expressiveness limitations of declarative visualization libraries \cite{Declarative}, we sought a solution capable of theoretically recreating any type of data visualization. Therefore, we directed the LLMs to utilize D3 \cite{bostock2011d3}, the most widely adopted imperative visualization programming library. This process expanded our dataset to quadruples $(I, D, V_h, V_{LLM})$, where each human-created visualization is paired with an LLM-generated counterpart.

To reflect real-world usage scenarios involving multi-turn dialogues and iterative refinement, we additionally recruited annotators to provide feedback on the generated visualizations. These annotators critiqued the initial LLM outputs and requested improved versions, simulating typical user-LLM interactions. We then incorporated those results of each iteration into our dataset. Since our work focuses on constructing programming language-agnostic, visual-oriented critic for visualizations, we filtered out instances where the generated code triggered compiler errors. For these cases, the most urgent form of critique is already provided by programming language-specific compiler error messages.

\subsection{Stage 4: High Quality Critique Collection}
The final stage involved the collection of critiques on LLM-generated visualizations. This extended our quadruples $(I, D, V_h, V_{LLM})$ to complete quintuplets $(I, D, V_h, V_{LLM}, C)$, where $C$ represents critique. We categorized reasonable visualization critiques into two main cases:

\begin{itemize}
\item Case 1: Critiques identifying defects that disqualify the visualization, highlighting specific issues that need correction.
\item Case 2: Critiques for visualizations without disqualifying defects, offering useful preferential suggestions (e.g., aesthetics, visual encoding effectiveness, etc.).
\end{itemize}

To enhance annotation quality and consistency, three visualization domain experts first conducted a preliminary experiment with 200 samples. They analyzed typical defects in LLM-generated visualizations, summarized them into categories, and provided examples with critique demonstrations for each category (Table \ref{tab:human_examples}). During the actual annotation process, we explicitly required annotators to methodically check for each defect type and provide corresponding feedback.

To assist annotators in providing preferential suggestions when no disqualifying defects were present, we used GPT-4o \cite{openai2024gpt4ocard} to automatically compare the LLM-generated visualization ($V_{LLM}$) with its corresponding human-created counterpart ($V_h$), generating potential preferential suggestions for annotators to reference. We recruited 10 annotators with visualization expertise and conducted annotation training. Each annotator's work was periodically sampled for quality assurance throughout the annotation process.

This approach yielded 2.7k high-quality critique annotations, with 0.16k reserved for subsequent testing purposes. The remaining 2.5k annotations were used for model training in subsequent phases.

\begin{figure}[t]
\fbox{\begin{minipage}{39em}
\scriptsize
\fontsize{8pt}{8pt}\selectfont
\textbf{Evaluation Instruction:}\\
Your task is to evaluate the candidate feedback given by a model for a generated visualization. You'll receive a "user instruction", a "generated visualization image", "ground truth feedback annotated by human", and "candidate feedback given by a model". Give a score based on the candidate feedback's quality.\\

5: When the visualization has defects, the feedback clearly highlights all of the defects. When the visualization has no defect, the feedback confirms that the visualization has no defects and provides very useful suggestions.\\
4: When the visualization has defects, the feedback confirms that the visualization has defects and points out those most significant defects. When the visualization has no defect, the feedback confirms that the visualization has no defect and provides useful suggestions.\\
3: When the visualization has defects, the feedback misses the significant defects but clearly confirms that the visualization has defects. When the visualization has no defect, the feedback confirms that the visualization has no defect and provides some less useful suggestions.\\
2: The feedback has a wrong judgment of the visualization (e.g., states defective visualization has no defects or states non-existent defects).\\
1: The feedback is completely wrong or contains severe hallucinations.\\

\{User Instruction, Generated visualization image, Ground truth feedback, Candidate feedback\}\\

\textbf{Example Evaluation Result by GPT-4o}:\\
The ground truth feedback identifies several defects in the visualization: the legend overlaps with the right side of the chart, the left part of the data marks exceeds the left end of the x-axis and overlaps with the y-axis, and the data marks corresponding to each tick label on the x-axis are difficult to distinguish, suggesting the addition of vertical gridlines for clarity. Upon examining the visualization, these defects are indeed present. However, the candidate feedback incorrectly states that the visualization has no defects, which is a wrong judgment. While the suggestion to improve visual hierarchy by adding more prominent borders or spacing between clusters is useful, the feedback fails to address the significant defects identified in the ground truth feedback. Therefore, the feedback is incorrect in its judgment and cannot score higher than 2.\\

Final Score: 2
\end{minipage}}
\caption{Instruction and scoring rubric for model-based automatic evaluation of visualization feedback. A 5-point Likert scale is used to assess critique quality, with detailed guidelines for comparing candidate feedback against human-annotated ground truth. An example evaluation result by GPT-4o is also provided to demonstrate the application of this scoring method.}
\label{fig:likert_instruction}
\end{figure}

\begin{figure}[t]
\fbox{\begin{minipage}{39em}
\scriptsize
\fontsize{8pt}{8pt}\selectfont
\textbf{Evaluation Instruction}: \\Two pieces of feedback have been provided for evaluating the same result of a visualization generation task. Pick the better feedback according
to whether it can identify defects when the generated visualization has defects that make it disqualified, or point out that there is no defect and provide useful suggestions when the
generated visualization has no defects but needs improvement. \\

\{Feedback 1, Feedback 2\}\\

Please choose from the following options. \\
A: Feedback 1 is better. \\
B: Feedback 2 is better. \\
C: Tie.
\end{minipage}}
\caption{ Instruction for human preference evaluation of visualization feedback. Evaluators are guided to compare two feedback examples and select the better one based on the accuracy of defect identification and suggestion quality for visualization improvement.}
\label{fig:pairwise_instruction}
\end{figure}

\section{The VIS-Shepherd Model}
We trained VIS-Shepherd with Qwen-2.5-VL-7B \cite{bai2025qwen2} as base model. The training was performed on the 2.5k critique annotations collected in the previous stage, allowing the model to learn from human expert feedback on visualization defects and improvement suggestions.

 For our training configuration, we employed the AdamW \cite{loshchilovdecoupled} optimizer with a learning rate of $1e-5 $ and a cosine learning rate scheduler. We set $\beta_{1}$=0.9, $\beta_{2}$=0.999, $\epsilon=1e-8 $ and weight decay of 0.01. The model was trained for just 1 epoch to prevent overfitting on our critique dataset. We used a batch size of 8 and a gradient accumulation step of 1. We used a maximum sequence length of 4096 tokens and applied a maximum gradient norm of 1.0 to ensure training stability. We set the number of warmup steps to 10 and utilized bf16 precision for computational efficiency. All computations were performed on 8 NVIDIA A800 (80GB) GPUs in an internal cluster, with each training run taking approximately 0.5 hours. In subsequent experiments, we fixed the temperature to 0 during model prediction to achieve most confident prediction results.

\section{Evaluating Feedback}
In this section, we describe our evaluation methodology for assessing VIS-Shepherd's performance, including our baseline selection, automated model-based evaluation approach, and human preference evaluation design.
\subsection{Baseline Models}
We compare Shepherd against the following state-of-the-art baselines:

\textbf{GPT-4o}: One of the most capable proprietary multimodal large language models developed by OpenAI. We included it to compare against leading commercial models.

\textbf{Llama-4-maverick}: One of the most capable open-source multimodal large language models developed by Meta, which is a 400B MoE model with 17B activated parameters. We selected it as a strong open-source baseline for visualization critique.

\textbf{Qwen-2.5-VL-7B}: The base model used for our training, pre-trained with 1 million synthetic chart-type samples, endowing it with chart understanding capabilities. This serves as our direct improvement baseline.

\textbf{Qwen-2.5-VL-72B}: It belongs to the same series as our base model Qwen-2.5-VL-7B, but with a parameter count 10 times larger. We included it to assess the impact of model scale on visualization critique performance.

\subsection{Model-based Automatic Evaluation}
\label{sec:Model-based Automatic Evaluation}
Leveraging LLMs to perform automated judging has been widely adopted across various complex evaluation scenarios \cite{surveyLLMasJudge}\cite{zheng2023judging}\cite{ wang2023shepherdcriticlanguagemodel}. We employed a similar approach by prompting GPT-4o to automatically evaluate the quality of feedback provided by different models against human critique annotations in our test set, using a 5-point Likert scale. We provided GPT-4o with the instructions shown in Figure \ref{fig:likert_instruction}, and directed it to analyze the critique based on these rating criteria in comparison with human annotations before providing a final score (see ``Example Evaluation Result by GPT-4o'' in Figure \ref{fig:likert_instruction}). We conducted a manual inspection of a series of output results to select and adjust our instructions until most evaluation results were consistent with human judgment.

\subsection{Human Preference Evaluation}
Due to potential biases or hallucinations in LLM evaluations \cite{trust}\cite{zheng2023judging}, we conducted a human preference evaluation to compare our model against several strong baselines. Human evaluators were provided with instructions as shown in Figure \ref{fig:pairwise_instruction} and were asked to select the superior feedback based on two criteria: the ability to identify disqualifying defects in generated visualizations, and the capacity to acknowledge when no disqualifying defects exist while offering constructive suggestions for improvement.

\section{Results}
In this section, we present evaluation results comparing VIS-Shepherd with baseline models across automated metrics, human preference evaluations, and ablation experiments.
\begin{figure}[t]
    \centering
    \includegraphics[width=1.0\linewidth]{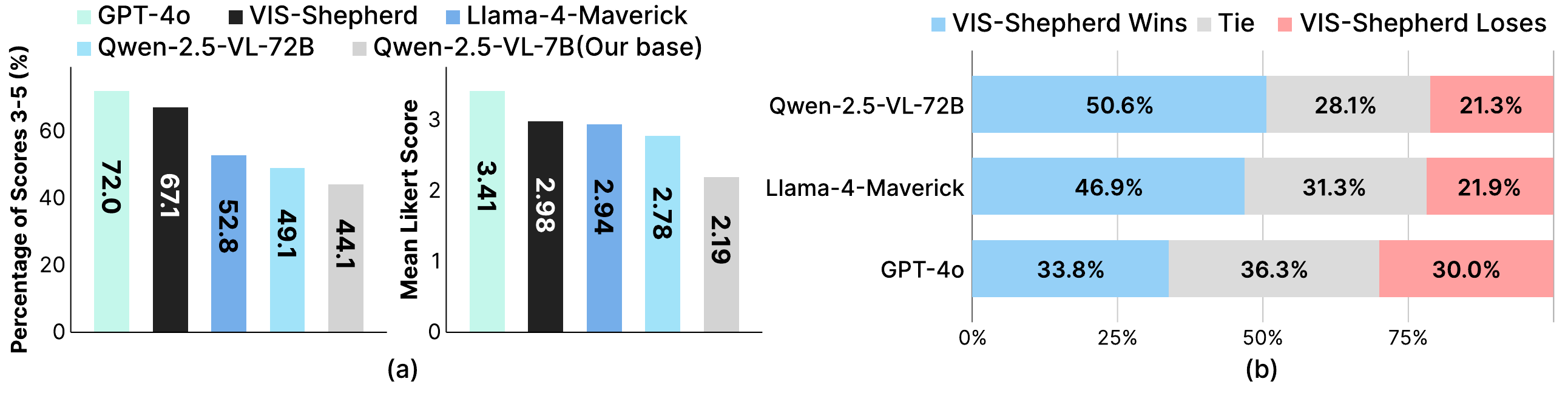}
    \caption{Figure \ref{fig:mainResult}: Comparison of VIS-Shepherd against baseline models. (a) Automated evaluation results showing percentage of high-quality responses (left) and mean Likert scores (right). (b) Human preference evaluation showing win-tie-loss rates in head-to-head comparisons. VIS-Shepherd outperforms strong open-source baselines and shows competitive performance against GPT-4o.}
    \label{fig:mainResult}
\end{figure}

\begin{figure}[t]
    \centering
    \includegraphics[width=1.0\linewidth]{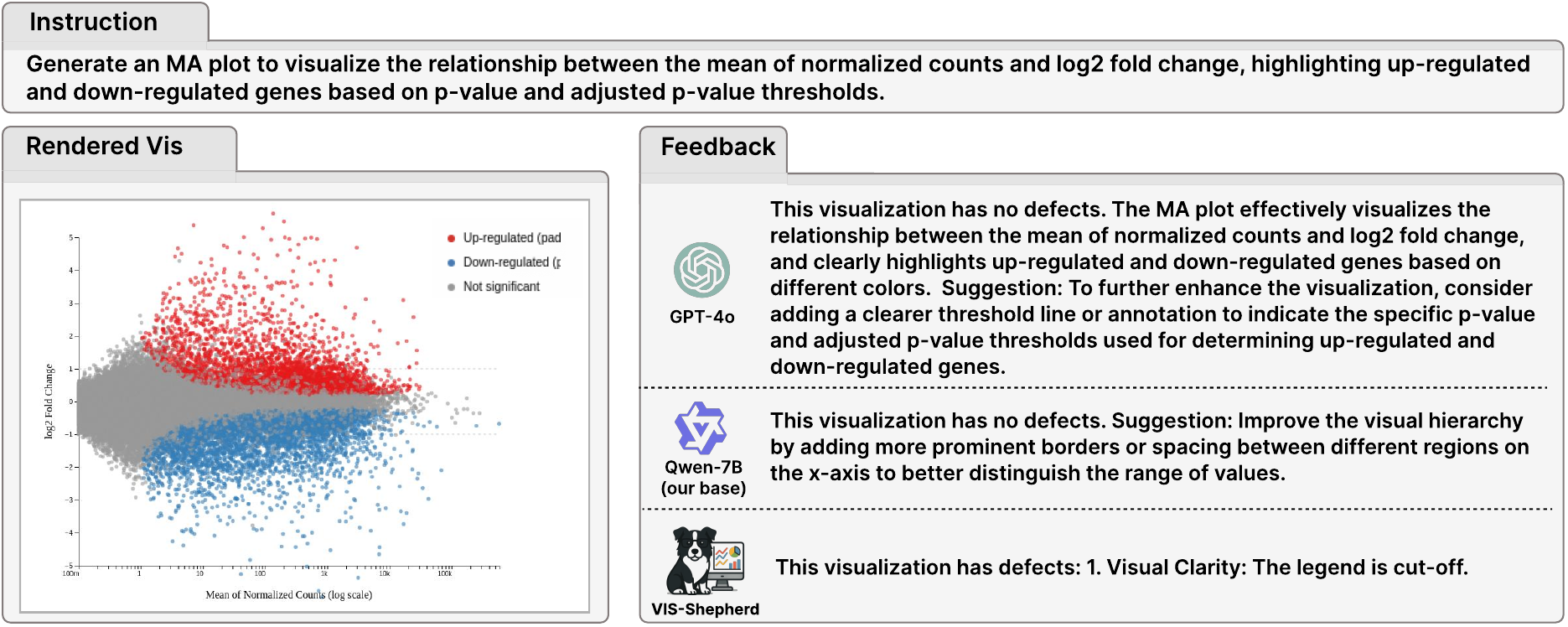}
    \caption{Qualitative comparison of model critiques for an MA plot. GPT-4o and Qwen-2.5-7B (our base model) miss the critical truncated legend issue, offering seemingly comprehensive critiques but nevertheless miss crucial visualization defects, while VIS-Shepherd correctly identifies this issue.}
    \label{fig:mainResult3}
\end{figure}

\subsection{Automated Evaluation Result}
We present a quantitative comparison of VIS-Shepherd against the baselines using model-based automated evaluation as described in Section~\ref{sec:Model-based Automatic Evaluation}. As shown in Figure \ref{fig:mainResult} (a), VIS-Shepherd shows improvement after training compared to its base model Qwen-2.5-VL-7B and outperforms Qwen-2.5-VL-72B, which belongs to the same model family but has ten times more parameters. Our results indicate that VIS-Shepherd's performance falls between the state-of-the-art open-source MLLM Llama-4-maverick and the proprietary model GPT-4o, performing better than the former but slightly below the latter. The score distribution analysis shows that both GPT-4o and VIS-Shepherd obtain more responses in the high-score range (3-5 on the Likert scale) compared to other models, suggesting the benefits of specialized training for visualization critique tasks.

\subsection{Human Preference Evaluation Result}
Following our automated evaluation, we conducted a human preference evaluation to validate our findings and better understand VIS-Shepherd's performance from human's perspective. Figure \ref{fig:mainResult} (b) presents the results of our head-to-head comparisons between VIS-Shepherd and the baseline models. Notably, these human preference results largely corroborate our automated evaluation findings: VIS-Shepherd demonstrates superior performance when compared against Qwen-2.5-VL-72B-Instruct and Llama-4-maverick with notably higher win rates, while achieving comparable performance with GPT-4o—a result that underscores the effectiveness of our training approach in producing critiques aligned with human judgment.

To further illustrate the qualitative improvements, Figure \ref{fig:mainResult3} provides a representative example that reveals important insights about model behavior. In this example, the visualization created based on user instructions contains a critical defect: the legend explaining the color meanings is truncated, significantly hampering interpretability. Interestingly, both GPT-4o and the original Qwen-2.5-VL-7B (our base model) failed to identify this fundamental issue, instead offering less urgent suggestions such as adding annotations or more prominent borders. In contrast, VIS-Shepherd successfully identifies and prioritizes this critical legend truncation problem, demonstrating the effectiveness of our specialized training approach in developing visualization-specific critique capabilities.

Furthermore, our analysis reveals an important methodological insight regarding visualization critique quality assessment: the verbosity of a critique does not necessarily correlate with its utility or accuracy. We observed instances where models produced lengthy, seemingly comprehensive critiques that nevertheless missed crucial visualization defects or offered misguided suggestions. This finding challenges the intuitive assumption that more detailed feedback is inherently superior, and suggests that evaluation metrics for visualization critique systems should prioritize the identification of critical issues over mere descriptive thoroughness. VIS-Shepherd, after training, demonstrates an improved ability to identify and prioritize fundamental visualization defects, even when they might not be immediately apparent to general-purpose models.

\subsection{Ablation Study}

\begin{wrapfigure}{R}{0.5\textwidth}
    \centering
    \vspace{-15pt}
    \includegraphics[width=0.48\textwidth]{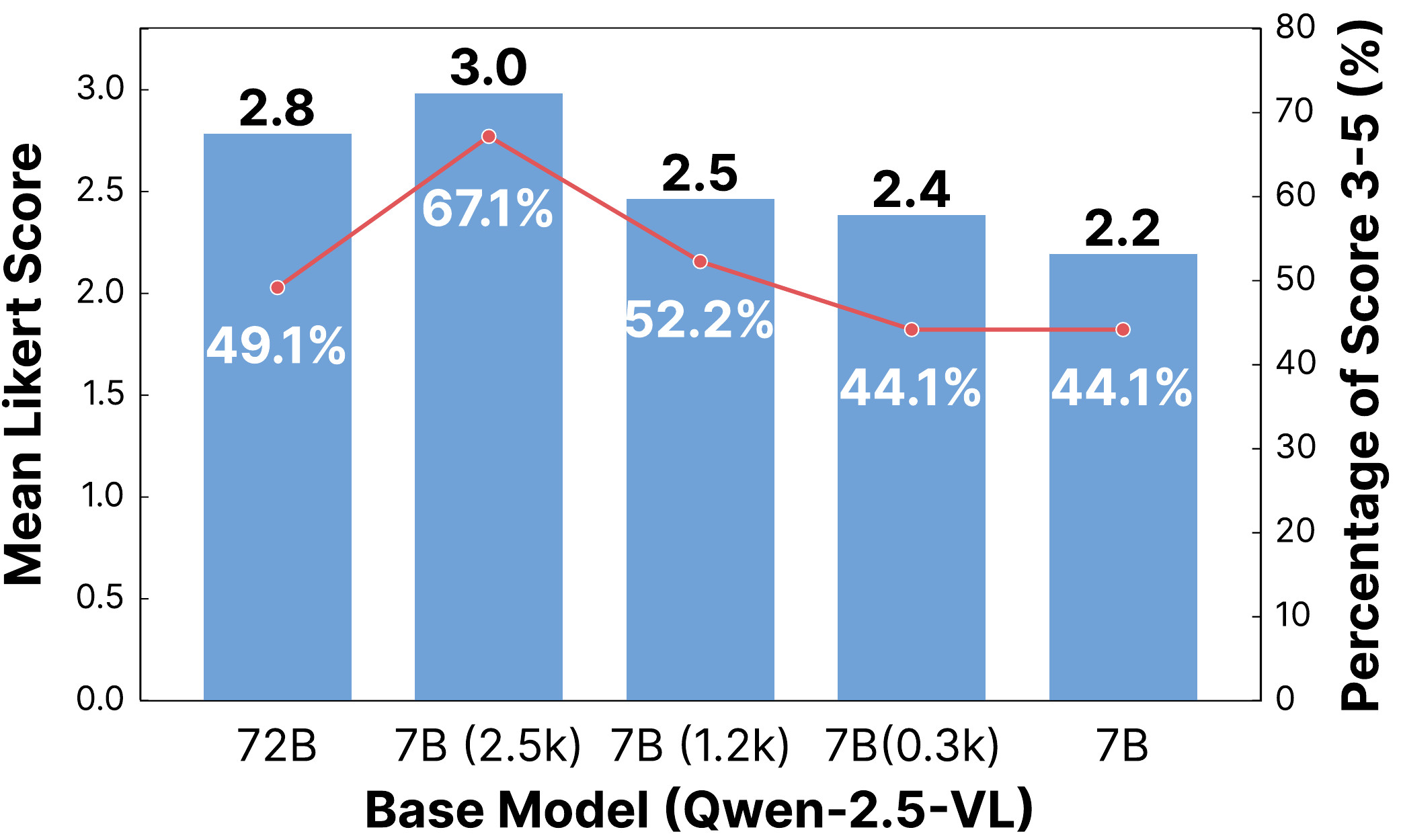} % 插入图片，宽度略小于 0.5\textwidth
    \caption{Ablation study on training data quantity. Performance improves with more high-quality examples, measured by mean Likert score (blue bar) and percentage of high-score responses (red line). }
    \label{fig:ablation}
\end{wrapfigure}

We conducted an ablation study to investigate how the quantity of high-quality training data influences model performance using our model-based automatic evaluation method. As shown in Figure \ref{fig:ablation}, VIS-Shepherd's critique capabilities show a general upward trend as we increase the volume of high-quality visualization critique examples in the training dataset (from 0 to 2.5k). The pattern culminates with the model eventually surpassing Qwen-2.5-VL-72B, a model from the same family with $\times 10$ parameter count. This finding highlights the effectiveness of our data-centric approach, suggesting that carefully curated domain-specific training data can be more impactful than simply scaling up model size for specialized visualization critique tasks.
\section{Limitations and Ethical Implications:}
Despite our efforts to construct a high-quality dataset of reasonable scale for our experiments, our approach would clearly benefit from larger-scale data and models. In future work, we plan to scale up our dataset and conduct experiments on larger models to further improve performance. Additionally, our current critic targets general visualization generation scenarios without optimization for specific visualization domains, which may limit its effectiveness in specialized domains. The critic-building pipeline and evaluation framework we've developed could be customized for specific domains if necessary. From an ethical perspective, we caution against over-reliance on automated feedback systems like ours. Users should maintain critical judgment when reviewing model-generated critiques, as these systems can inherit biases from training data and may occasionally miss important context-specific considerations that human experts would catch.

\section{Conclusion}
In this paper, we introduced VIS-Shepherd, a specialized MLLM-based critic for evaluating and providing feedback on LLM-generated data visualizations. Our framework for constructing high-quality visualization critique datasets enabled even small (7B) models to achieve performance comparable to much larger counterparts, demonstrating that domain-specific expertise embedded in well-curated datasets can effectively compensate for model scale. Through comprehensive evaluations, we showed that our approach makes significant progress toward improving LLM-generated visualizations by providing targeted feedback on visualization-specific defects. The findings suggest that dataset quality and domain focus are crucial elements for developing visualization critics, highlighting potential directions for visualization evaluation approaches. As we continue to advance automated visualization critique, future research could explore more ways to leverage visualization critic models to enhance the quality of visualization generation, and domain-specific adaptations to further enhance the capabilities of visualization critics.

% \begin{ack}
% Use unnumbered first level headings for the acknowledgments. All acknowledgments
% go at the end of the paper before the list of references. Moreover, you are required to declare
% funding (financial activities supporting the submitted work) and competing interests (related financial activities outside the submitted work).
% More information about this disclosure can be found at: \url{https://neurips.cc/Conferences/2025/PaperInformation/FundingDisclosure}.

% Do {\bf not} include this section in the anonymized submission, only in the final paper. You can use the \texttt{ack} environment provided in the style file to automatically hide this section in the anonymized submission.
% \end{ack}

% \section*{References}
\medskip

\bibliographystyle{unsrt}  
\bibliography{ref}

%%%%%%%%%%%%%%%%%%%%%%%%%%%%%%%%%%%%%%%%%%%%%%%%%%%%%%%%%%%%
\clearpage

\appendix
\section{Appendix}

\subsection{Source Code and Dataset}
Please check the dataset and the source code for training and evaluating VIS-Shepherd in our anonymous repository \url{ https://github.com/bopan3/VIS-Shepherd}. 
\subsection{Dataset Construction Details}
We provide more details about each stage during our dataset construction process in this section.

\subsubsection{Stage 1: Human-created Instance Curation}
\textbf{Automatic extraction of visualization instances}:
Observable\footnote{https://observablehq.com/} allows users to host their visualization code in notebook format. To automatically extract individual visualization instances, we implemented an automatic visualization code extraction pipeline for each notebook. We first automatically detect code cells containing any rendered outputs. We then trace backward to identify all cells on which the rendering cell depends, and automatically concatenate them in dependency order to form a complete visualization code segment. Each of these self-contained segments constitutes a candidate visualization instance for downstream filtering.

\textbf{Automatic filtering with MLLM}: We employed Gemini-2.0-flash \cite{gemini2023} as the MLLM to filter out instances where "the rendered content is not a visualization," "the quality of the visualization instance is low," or "the visualization contains dynamic elements." This filtering process reduced the manual curation workload in subsequent steps. The prompt used in this process is presented in \textit{``Prompt 1: Prompt for Automatic Filtering with MLLM''} as follows.

\begin{tcolorbox}[breakable, colback=white, title={Prompt1: Prompt for Automatic Filtering with MLLM}]
\textbf{System Prompt:}\\
\#\# Instruction \\
You are an intelligent visualization analyzer. I will give you "the source code of a html file", "the rendering result of the html file", and "criteria for filtering", then you should determine whether to filter out this html file. \\
Your output must exactly follow the Output Format Specification, and you must not add any other text or comments.\\

\#\# Provided Inputs\\
Input1. The source code of a html file is given as follows:\\
\{html\_code\}\\

Input2. The rendering result of the html file is shown as the attached image.\\

Input3. The criteria for filtering is given as follows:\\\\
Step1: \\
Description: check if this html file is a data visualization (must map a given dataset to a visual elements), if not, filter it out.\\
Label: if filter out at this step, the label is "Not Data Visualization"\\
Step2: check if the rendering result is of low quality (e.g. is blank / almost blank, with illegible/overlapping text/elements, of low readability, of severely incomplete representations, using toy datasets etc.), if yes, filter it out.\\
Label: if filter out at this step, the label is "Low Quality Visualization"\\
Step3: check if the html file contains any animation or interaction, if yes, filter it out.\\
Label: if filter out at this step, the label is "Not Static Visualization"\\

\#\# Output Format Specification:\\
\lbrack THINKING\rbrack\\
At here, output your thinking process for filtering.\\
The format must be exactly as follows:\\
\texttt{\`}\texttt{\`}\texttt{\`} markdown\\
...\\
\texttt{\`}\texttt{\`}\texttt{\`} \\
\lbrack \slash THINKING\rbrack\\
\lbrack FILTERING\_RESULT\rbrack\\
At here, output the filtering result.\\
The format must be exactly as follows:\\
\texttt{\`}\texttt{\`}\texttt{\`} json\\
\{\{\\
    "Filtered": true/false,\\
    "Label": "xx"\\
\}\}\\
\texttt{\`}\texttt{\`}\texttt{\`} \\
\lbrack \slash FILTERING\_RESULT\rbrack
\end{tcolorbox}

\begin{figure*}[]
        \centering
        \includegraphics[width=1.0\textwidth]{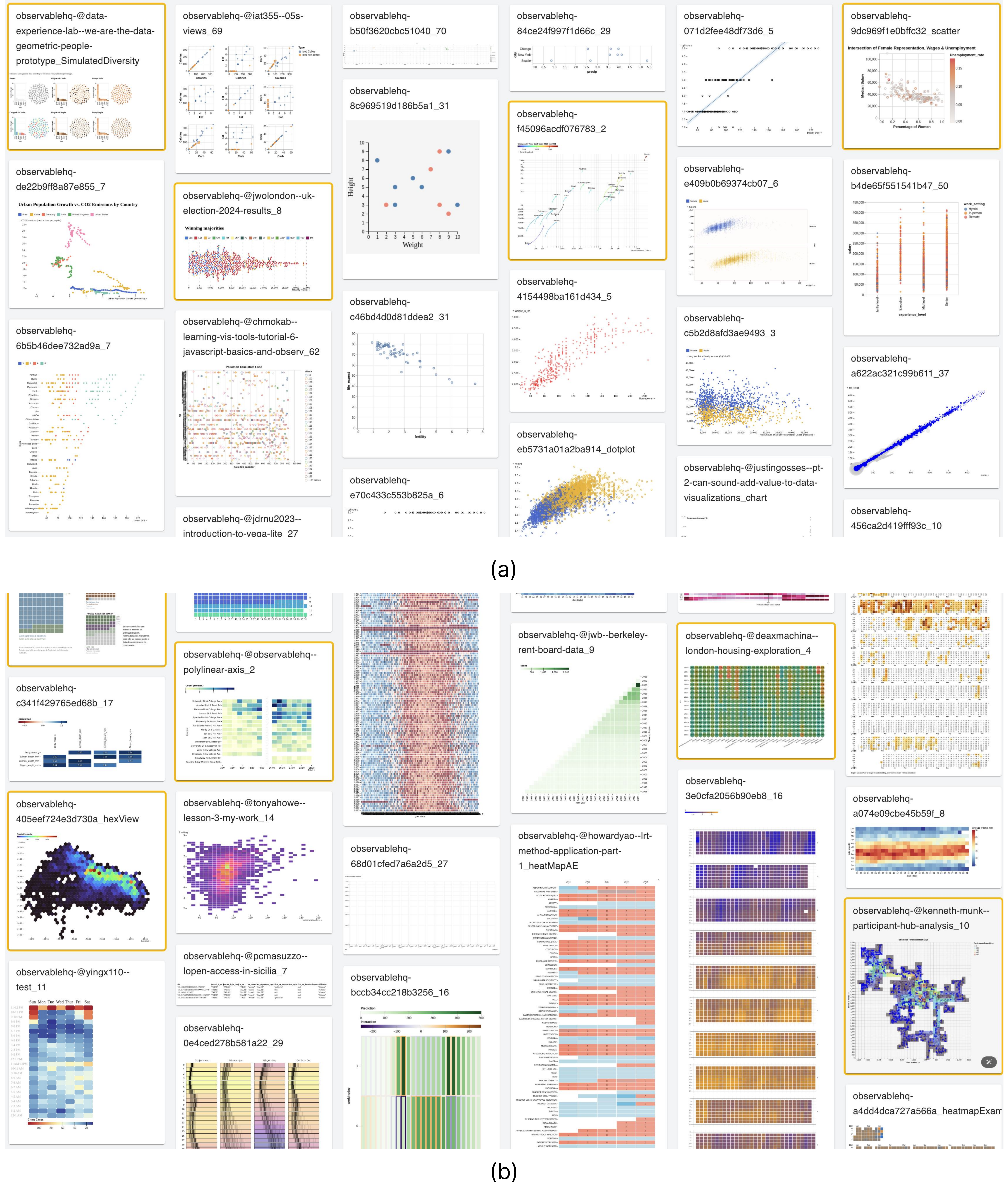}
    \caption{Screenshots of the annotation system for manual high-quality visualization selection. (a) Candidate visualizations of the "point" type and (b) candidate visualizations of the "grid" type. Visualizations selected by annotators are highlighted with yellow borders.}
    \label{fig:manual_selection}
\end{figure*}

\textbf{Manual high-quality instance selection}: We developed an online annotation system to assist annotators in selecting high-quality visualization instances. To encourage diversity in the final selection and support annotators' decision-making, we pre-classified all visualizations using Gemini-2.0-flash \cite{gemini2023} based on an image-based typology for visualization \cite{chen2025imagebasedtypologyvisualization}. The prompt used for this classification is presented in \textit{``Prompt 2: Prompt for Visualization Classification with MLLM''}. During each annotation round, annotators were presented with 50 visualizations belonging to the same category and asked to select those of relatively high quality. Figure~\ref{fig:manual_selection} shows two representative screenshots of the annotation system: screenshot (a) displays a set of candidate visualizations of the ``point'' type, while screenshot (b) shows candidate visualizations of the ``grid'' type. The visualization instances selected by the annotators will be highlighted with a yellow border.

\begin{tcolorbox}[breakable, colback=white, title={Prompt2: Prompt for Visualization Classification with MLLM}]
\textbf{System Prompt:}\\
\#\# Instruction \\
You are an intelligent visualization analyzer. I will give you "the image of data visualization", and "candidate types for classification", then you should classify this visualization based on the given candidate types.\\
Your output must exactly follow the Output Format Specification, and you must not add any other text or comments.\\

\#\# Provided Inputs\\
Input1. The image of data visualization is shown as the attached image.\\

Input2. The candidate types for classification is given as follows:\\

1. Generalized Bar Representations\\
Label: Bar\\
Description: Graphs that represent data with straight bars that can be arranged on a straight or curved baseline and whose heights or lengths are proportional to the values they represent.\\
Examples: bar charts, stacked bar charts, box plots, or sunburst diagrams.\\
2. Point-based Representations \\
Label: Point\\
Description: Representations that use point locations. These locations are often shown using dots or circles, but also other shapes such as 3D spheres, triangles, stars, etc.\\
Examples: scatterplots, point clouds, dot plots, or bubble charts.\\
3. Line-based Representations\\
Label: Line\\
Description: Representations where information is emphasized through (straight or curved) lines.\\
Examples: line charts, parallel coordinates, contour lines, radar/spider charts, streamlines, or tensor field lines.\\
4. Node-link Trees/Graphs, Networks, Meshes\\
Label: Node-link\\
Description: Representations using points for and explicit connections between these points\\
to convey relationships between data values.\\
Examples: node-link diagrams, node-link trees, node-link graphs, meshes, arc diagrams, or Sankey diagrams.\\
5. Area-based Representations\\
Label: Area\\
Description: Representations with a focus on areas of 2D space or 2D surfaces including sub-sets of these surfaces. Areas can be geographical regions or polygons whose size or shape represents abstract data.\\
Examples: (stacked) area chart, streamgraph, ThemeRiver, violin plot, cartograms, histograms, ridgeline chart, Voronoi diagram, treemaps, pie chart.\\
6. Generalized Matrix / Grid\\
Label: Grid\\
Description: Representations that separate data into a discrete spatial grid structure. The grid often has rectangular cells but may also use other shapes such as hexagons or cubes. Elements such as glyphs or a color encoding can appear in the grid cells.\\
Examples: network matrices, discrete heatmaps, scarf or strip plots, space-time cubes, or matrix-based network visualizations.\\
7. Continuous Color and Grey-scale, and Textures\\
Label: Continuous-ColorPattern\\
Description: Representations of structured patterns across an image or atop a geometric 3D object. These patterns can be evoked by changes in intensity, changes in hue, brightness, and/or saturation. The changes are typically smooth (continuous) but could show sharp transitions as well.\\
Examples: Directional patterns such as Line Integral Convolution (LIC), Spot Noise, and Image-Space Advection (ISA) to show flow fields, continuous heatmaps, intensity fields, or even a binary image.\\
8. Glyph-based Representations\\
Label: Glyph\\
Description: Multiple small independent visual representations (often encoded by position and additional dimensions using color, shape, or other geometric primitives) that depict multiple attributes (dimensions) of a data record. Placement is usually meaningful and typically multiple glyphs are displayed for comparison.\\
Examples: Star glyphs, 3D glyphs, Chernoff faces, vector field glyphs.\\
9. Text-based Representations\\
Label: Text\\
Description: Representations of data (often text itself) that use varying properties of letters/words such as font size, color, width, style, or type to encode data.\\
Examples: Tag clouds, word trees, parallel tag clouds, typomaps.\\

\#\# Output Format Specification:\\
\lbrack THINKING\rbrack\\
At here, output your thinking process for classification.\\
The format must be exactly as follows:\\
\texttt{\`}\texttt{\`}\texttt{\`} markdown\\
...\\
\texttt{\`}\texttt{\`}\texttt{\`} \\
\lbrack \slash THINKING\rbrack\\
\lbrack CLASSIFICATION\_RESULT\rbrack\\
At here, output the classification result.\\
Note that the label must be the label of one of the candidate types for classification.\\
The format must be exactly as follows:\\
\texttt{\`}\texttt{\`}\texttt{\`} json\\
\{\{\\
    "Label": "xx"\\
\}\}\\
\texttt{\`}\texttt{\`}\texttt{\`} \\
\lbrack \slash CLASSIFICATION\_RESULT\rbrack
\end{tcolorbox}

\subsubsection{Stage 2: Instruction Synthesis and Dataset Exportation}
\textbf{Instruction Synthesis}: We utilized Claude-3.5-Sonnet \cite{anthropic2024claude} to synthesize natural language instructions for each visualization instance. The model was prompted to produce a list of candidate queries that could plausibly result in the given visualization. To encourage diversity, we ask the MLLM to first brainstorm contextual information such as ``the profile of the user who provides the query'', ``the level of data visualization expertise of the user'', and ``the usage scenario of this visualization image'' before synthesizing the final instructions. The detailed prompt can be found in \textit{``Prompt 3: Prompt for Instruction Synthesis''} as follows.

\begin{tcolorbox}[breakable, colback=white, title={Prompt3: Prompt for Instruction Synthesis}]
\textbf{System Prompt:}\\
\#\# Instruction \\
You are an intelligent programmer specialized in visualization. I will give you "a visualization image", "the code that generates the visualization image", "the illustration of the data files used for generating the visualization image", and you should generate a query that can be used to generate this visualization image:\\
Your output must exactly follow the Output Format Specification, and you must not add any other text or comments.\\

\#\#\# Provided Inputs\\
\#\#\#\# Input1. The visualization image is given as the attached image.\\

\#\#\#\# Input2. The code that generates the visualization image is given as follows:\\
\{vis\_code\}\\

\#\#\#\# Input3. The illustration of the data files used for generating the visualization image is given as follows:\\
\{data\_illustration\}\\

\#\#\# Output Format Specification:\\
\lbrack SUBSECTION1\_RESULT\_LIST\rbrack\\
Generate a list of candidate queries that can be used to generate this visualization image.\\
Before you generate each query, you need to first generate "the profile of the user who provides the query", "the level of data visualization expertise of the user", and "the usage scenario of this visualization image".\\
The query should be a natural language query that can be used to generate the visualization image, which could be either a visualization design utterance or a data analysis question.\\
The format must be exactly as follows:\\
\texttt{\`}\texttt{\`}\texttt{\`} json\\
\lbrack\\
\{\{"user\_profile": "...", "data\_vis\_expertise": "Low/Medium/High", "usage\_scenario": "...", "query": "..."\}\},\\
\{\{"user\_profile": "...", "data\_vis\_expertise": "Low/Medium/High", "usage\_scenario": "...", "query": "..."\}\},\\
...\\
\rbrack\\
\texttt{\`}\texttt{\`}\texttt{\`}\\
\lbrack \slash SUBSECTION1\_RESULT\_LIST\rbrack
\end{tcolorbox}

\textbf{Dataset Exportation}: To perform uniform dataset exportation for in-the-wild visualization instances, we had Claude-3.5-Sonnet~\cite{anthropic2024claude} generate customized dataset exportation scripts for each visualization instance using the prompt shown in \textit{``Prompt4: Prompt for Dataset Exportation''}. To verify that the exported dataset was complete and usable, we then had Claude-3.5-Sonnet~\cite{anthropic2024claude} rewrite a version of the visualization code that achieved the same visual effect but read from the exported data, using the prompt shown in \textit{``Prompt5: Prompt for Verifying Exported Dataset''}. Similar to previous work \cite{lida, VisEval}, for each type of exported dataset, we generated a rule-based dataset preview to help the LLM understand the dataset, with examples shown in Figure \ref{fig:dataset_preview}.

\begin{tcolorbox}[breakable, colback=white, title={Prompt4: Prompt for Dataset Exportation}]
\textbf{System Prompt:}\\
\#\# Instruction \\
You are an intelligent programmer specialized in visualization. I will give you "an html file about visualization", and you should add additional data export logic in the original html file such that all data that are necessary for the visualization are exported in suitable formats such that they can be easily used by other people to reproduce this visualization.\\

\#\# Provided Inputs\\
Input1. The html file about visualization is given as follows:\\
\{html\_file\}\\

\#\# Output format specification:\\
\lbrack SUBSECTION1\_PLAN\rbrack\\
At here, analyze the html file and plan for the output data formats such that all data that are necessary for the visualization are exported in suitable formats such that they can be easily used by other people to reproduce this visualization.\\
Note that you should just export the raw source data, not the derived data (e.g. no need to export the final rendering result, no need to export derived data from the raw source data).\\
Note that currently we only support the following file types: \\
csv\_table - a csv file that contains the table data for the visualization\\
json\_topojson - a json file that contains the topology data with specialized topojson format for the visualization\\
json\_geojson - a json file that contains the geojson data for the visualization\\
png\_image - a png image file that contains certain image resource\\
jpg\_image - a jpg image file that contains certain image resource\\
svg\_image - a svg image file that contains certain image resource\\
If you find data in other unsupported formats (e.g. normal json file), you must convert it to csv\_table format (the tidy data theory makes sure it's possible).\\
The format must be exactly as follows:\\
\texttt{\`}\texttt{\`}\texttt{\`} json\\
\{\{\\
    "file\_list": \lbrack\\
        \{\{\\
            "file\_name": "file\_name\_1.csv",\\
            "file\_type": "csv\_table",\\
            "description": "provide a description to help other people understand the content of this file"\\
        \}\},\\
        \{\{\\
            "file\_name": "file\_name\_2.json",\\
            "file\_type": "json\_topojson",\\
            "description": "provide a description to help other people understand the content of this file"\\
        \}\},\\
        \{\{\\
            "file\_name": "file\_name\_3.png",\\
            "file\_type": "png\_image",\\
            "description": "provide a description to help other people understand the content of this file"\\
        \}\}\\
    \rbrack\\
\}\}\\
\texttt{\`}\texttt{\`}\texttt{\`}\\
\lbrack \slash SUBSECTION1\_PLAN\rbrack\\
\lbrack SUBSECTION2\_EDITED\_CODE\rbrack\\
As here, add additional data export logic in the original html file such that we can export all the files planned in the previous section.\\
Note that you should output complete html code WITHOUT any omission. \\
Note that we do exportion by pushing the encoded data to the globalThis.exported\_data array. Don't use any other way to do the exportion. Don't add any button or other UI elements to the html file.\\
Example code snippet:\\
// EXPORT\_LOGIC\_FOR\_FILE\_NAME\_1\_START\\
    const csvContent = some\_customed\_functions\_or\_processes();\\
    const csvEncoded = btoa(csvContent);\\
    globalThis.exported\_data = globalThis.exported\_data || \lbrack\rbrack;\\
    globalThis.exported\_data.push(\{\{\\
        filename: "file\_name\_1.csv",  // make sure the file name is the same as the file\_name\_1 in the previous section\\
        data: csvEncoded,\\
        type: "text/csv"\\
    \}\});\\
// EXPORT\_LOGIC\_FOR\_FILE\_NAME\_1\_END\\
The format must be exactly as follows:\\
\texttt{\`}\texttt{\`}\texttt{\`} html\\
...\\
\texttt{\`}\texttt{\`}\texttt{\`}\\
\lbrack \slash SUBSECTION2\_EDITED\_CODE\rbrack
\end{tcolorbox}
\begin{tcolorbox}[breakable, colback=white, title={Prompt5: Prompt for Verifying Exported Dataset}]
\textbf{System Prompt:}\\
\#\# Instruction \\
You are an intelligent programmer. I will provide you with "reference html file for visualization", "the illustration for the multiple source data files under location '.\slash\{input\_data\_folder\_name\}' ", and you should make minimal edit to the reference html file such that it switch to read the multiple source data files and visualize them.\\

\#\# Provided Inputs\\
Input1. reference html file for visualization:\\
\{reference\_html\_file\_for\_visualization\}\\

Input2. the illustration for the multiple source data files under location '.\slash\{input\_data\_folder\_name\}' :\\
\{multi\_src\_data\_illustration\}\\

\#\# Output Format Specification:\\
\lbrack EDITED\_CODE\rbrack\\
As here, make the minimal edit to the reference html file such that it switch to read the multiple source data files and visualize them.\\
Note that you should read the multiple source data files (under location '.\slash\{input\_data\_folder\_name\}') directly, no need to use FileAttachment function.\\
Note that you should output complete html code.\\
The format must be exactly as follows:\\
\texttt{\`}\texttt{\`}\texttt{\`} html\\
...\\
\texttt{\`}\texttt{\`}\texttt{\`}\\
\lbrack \slash EDITED\_CODE\rbrack
\end{tcolorbox}

\begin{figure}[t]
\begin{minipage}{39em}

\begin{lstlisting}[style=json, caption={}]
// CSV Table Preview Example:
[
    {"column": "sex", "properties": {"dtype": "category", "samples": ["female", "male"],
    "num_unique_values": 2}}, 

    {"column": "rank", "properties": {"dtype": "category", "samples": ["professor", "lecture",
    "assistant professor"], "num_unique_values": 4}}, 

    {"column": "salary", "properties": {"dtype": "number", "std": 2902, "min": 778, "max": 9684,
    "samples": [2545, 3432, 1299], "num_unique_values": 13}}
]
// GeoJSON Preview Example:
{
    "type": "FeatureCollection",
    "feature_count": 50,
    "geometry_types": ["Point", "Polygon"],
    "property_fields": ["name", "population", "area"],
    "feature_samples": [
        {
            "type": "Feature",
            "geometry_type": "Point",
            "properties": {"name": "New York", "population": 8804190, "area": 783.8}
        },
        {
            "type": "Feature",
            "geometry_type": "Polygon",
            "properties": {"name": "Los Angeles", "population": 3898747, "area": 1302}
        },
        {
            "type": "Feature", 
            "geometry_type": "Point",
            "properties": {"name": "Chicago", "population": 2746388, "area": 606.1}
        }
    ],
    "bbox": [-180, -90, 180, 90],
    "coordinate_stats": {
        "lat": {"min": -90, "max": 90, "mean": 12.5},
        "lng": {"min": -180, "max": 180, "mean": 25.3}
    },
    "file_size_kb": 325.7
}
\end{lstlisting}

\end{minipage}
\caption{Examples of rule-based dataset previews used to help the LLM understand the data structure. The figure shows a CSV table preview and a GeoJSON preview with sample data and metadata.}
\label{fig:dataset_preview}
\end{figure}

\subsubsection{Stage 3: LLM-based Visualization Generation}

To capture the spectrum of visualization defects typically produced by state-of-the-art LLMs, we utilized advanced models like GPT-4o \cite{openai2024gpt4ocard} and Claude-Sonnet-3.5 \cite{anthropic2024claude} to generate visualizations based on the instruction-dataset pairs from Stage 2. The detail of the prompt used for generation is shown in \textit{``Prompt6: Prompt for Visualization Generation''}. To reflect real-world usage scenarios involving multi-turn dialogues and iterative refinement, we additionally recruited annotators to provide feedback on the generated visualizations. These annotators critiqued the initial LLM outputs and requested improved versions, simulating typical user-LLM interactions. The prompt for LLM to improve visualization based on user feedback is shown in \textit{``Prompt7: Prompt for Visualization Improvement Based on Feedback''}.

\begin{tcolorbox}[breakable, colback=white, title={Prompt6: Prompt for Visualization Generation }]
\textbf{System Prompt:}\\
\#\# Instruction \\
You are a data visualization expert. I will provide you with "the illustration of the data files used for generating the visualization image (those files are under the path '.\slash data\_folder')", "the user's query", and you should write visualization code (using d3.js v7) that fulfill "the user's query" as closely as possible.\\
Your output must exactly follow the Output Format Specification, and you must not add any other text or comments.\\

\#\# Provided Inputs\\
\#\#\# Input1. The illustration of the data files used for generating the visualization image is given as follows:\\
\{data\_illustration\}\\

\#\#\# Input2. The user's query is given as follows:\\
\{user\_query\}\\

\#\# Output Format Specification:\\
\lbrack THINKING\rbrack\\
At here, output your thinking process.\\
The format must be exactly as follows:\\
\texttt{\`}\texttt{\`}\texttt{\`} text\\
...\\
\texttt{\`}\texttt{\`}\texttt{\`}\\
\lbrack \slash THINKING\rbrack\\
\lbrack CODE\rbrack\\
At here, output the visualization code.\\
The format must be exactly as follows:\\
\texttt{\`}\texttt{\`}\texttt{\`} html\\
...\\
\texttt{\`}\texttt{\`}\texttt{\`}\\
\lbrack \slash CODE\rbrack
\end{tcolorbox}

\begin{tcolorbox}[breakable, colback=white, title={Prompt7: Prompt for Visualization Improvement Based on Feedback}]
\textbf{System Prompt:}\\
\#\# Instruction \\
You are a data visualization expert. Below is the code you provided to meet my Original Instruction.\\
\{previous\_code\}\\

Please try your best to improve the code based on the following feedback for improvement:\\
\{feedback\_for\_improvement\}\\

Your output must exactly follow the format as follows, and you must not add any other text or comments:\\
\lbrack THINKING\rbrack\\
At here, output your thinking process. \\
Note that you should focus on making specific modifications based on the feedback for improvement (less is more).\\
Note that you should never add any interaction or animation or print statement to the code.\\
Note that you should not add comments to explain where the improvement is made.\\
The format must be exactly as follows:\\
\texttt{\`}\texttt{\`}\texttt{\`} text\\
...\\
\texttt{\`}\texttt{\`}\texttt{\`}\\
\lbrack \slash THINKING\rbrack\\
\lbrack IMPROVED\_CODE\rbrack\\
\texttt{\`}\texttt{\`}\texttt{\`} html\\
...\\
\texttt{\`}\texttt{\`}\texttt{\`}\\
\lbrack \slash IMPROVED\_CODE\rbrack\\

\#\# Original Instruction\\
You are a data visualization expert. I will provide you with "the illustration of the data files used for generating the visualization image (those files are under the path '.\slash data\_folder')", "the user's query", and you should write visualization code (using d3.js v7) that fulfill "the user's query" as closely as possible.\\
Your output must exactly follow the Output Format Specification, and you must not add any other text or comments.\\

\#\#\# Provided Inputs\\
\#\#\#\# Input1. The illustration of the data files used for generating the visualization image is given as follows:\\
\{data\_illustration\}\\

\#\#\#\# Input2. The user's query is given as follows:\\
\{user\_query\}\\

\#\#\# Output Format Specification:\\
\lbrack THINKING\rbrack\\
At here, output your thinking process.\\
The format must be exactly as follows:\\
\texttt{\`}\texttt{\`}\texttt{\`} text\\
...\\
\texttt{\`}\texttt{\`}\texttt{\`}\\
\lbrack \slash THINKING\rbrack\\
\lbrack CODE\rbrack\\
At here, output the visualization code.\\
The format must be exactly as follows:\\
\texttt{\`}\texttt{\`}\texttt{\`} html\\
...\\
\texttt{\`}\texttt{\`}\texttt{\`}\\
\lbrack \slash CODE\rbrack
\end{tcolorbox}

\subsubsection{Stage 4: High Quality Critique Collection}
\begin{figure*}[]
        \centering
        \includegraphics[width=1.0\textwidth]{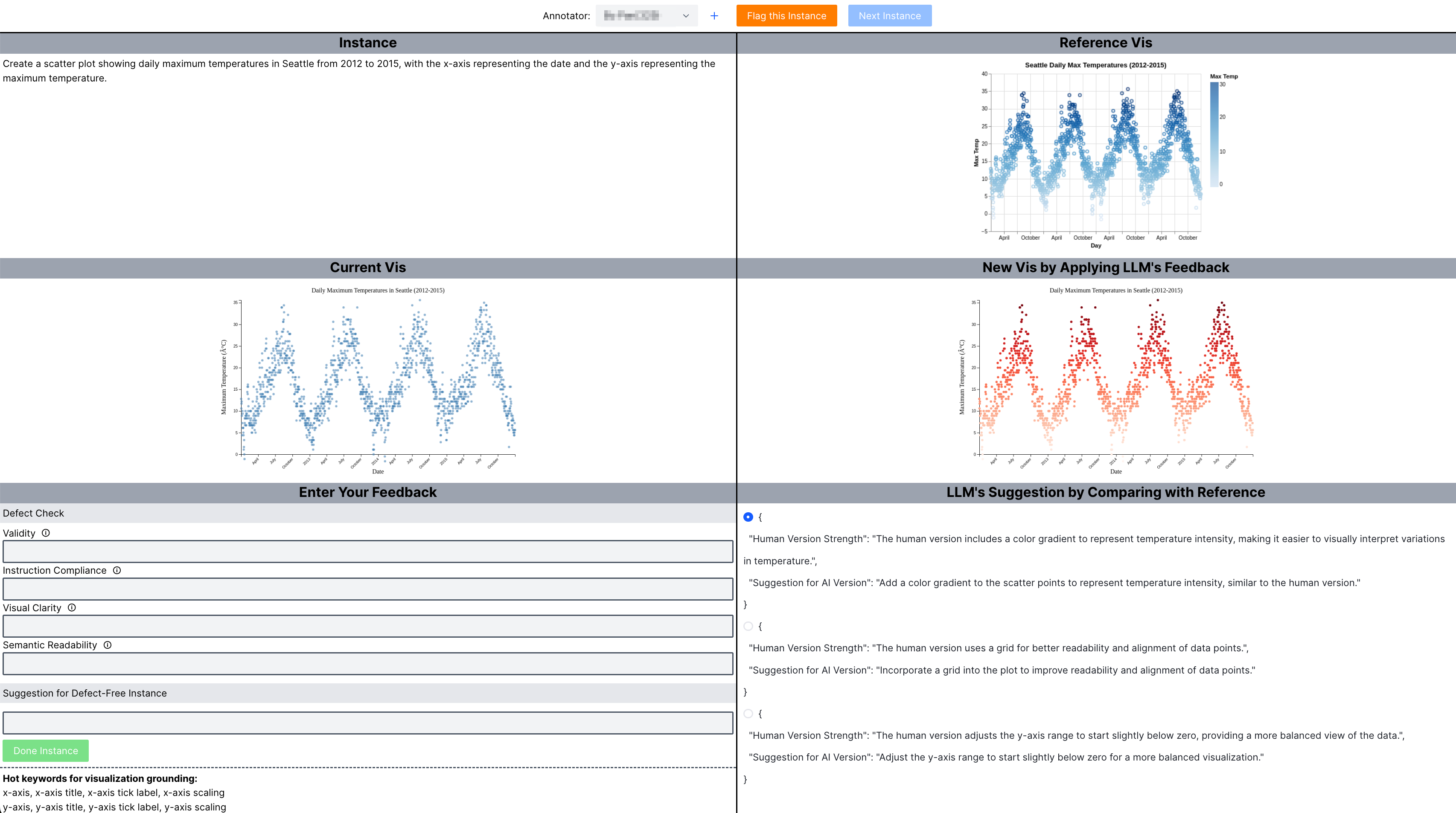}
    \caption{Screenshot of the annotation system for critique collection. The interface displays the instruction for current instance (top left), the LLM-generated visualization based on the instruction (middle left), a human-created visualization for reference (top right), an area for annotator to provide feedback (bottom left), automatically generated suggestions for annotator's reference (bottom right), and the improved versions based on automatically generated suggestions for annotator's reference (middle right).}
    \label{fig:critic_annotation}
\end{figure*}

We developed an online annotation system to assist annotators in providing critique for LLM-generated visualizations. Figure~\ref{fig:critic_annotation} provides a screenshot of the annotation system. Based on the instruction for current instance (top left), the current LLM-generated visualization based on the instruction (middle left), and the reference visualization created by human (top right), the annotators provide the critique for the current LLM-generated visualization (bottom left). The annotators need to first check if the visualization contains any defects and enter critique in the corresponding cell if they identified a type of defect. If there is no defect, the annotators enter the suggestion for further improvement in the cell named ``Suggestion for Defect-free Instance''. To facilitate annotation when no disqualifying defects were present, we used GPT-4o \cite{openai2024gpt4ocard} to automatically compare the LLM-generated visualization with the human-created counterpart to generate several reference suggestions for annotators (bottom right, the detailed prompt for suggestion generation is shown in \textit{``Prompt8: Prompt for Reference Suggestion Generation''}). We also let Claude-Sonnet-3.5 \cite{anthropic2024claude} improve the current LLM-generated visualization based on the feedback and show the improved result for more intuitive reference (middle right).

\begin{tcolorbox}[breakable, colback=white, title={Prompt8: Prompt for Reference Suggestion Generation}]
\textbf{System Prompt:}\\
\#\# Instruction \\
You are a data visualization expert. I will provide you with "the user's query for generating the visualization", "the current visualization generated by AI" and "the reference visualization created by human". You should provide \{num\_of\_suggestion\} most important suggestions for "the current visualization generated by AI". You can refer to "the reference visualization created by human".\\
Note that we want a static visualization, so do not suggest any interaction or animation.\\
Your output must exactly follow the Output Format Specification, and you must not add any other text or comments.\\

\#\# Provided Inputs\\
\#\#\# Input1. The user's query for generating the visualization is given as follows:\\
\{user\_query\}\\

\#\#\# Input2. The current visualization generated by AI is shown in the first attached image.\\

\#\#\# Input3. The reference visualization created by human is shown in the second attached image.\\

\#\# Output Format Specification:\\
\texttt{\`}\texttt{\`}\texttt{\`} json\\
\lbrack\\
    \{\{\\
        "Human Version Strength": "xx",\\
        "Suggestion for AI Version": "xx",\\
    \}\},\\
    \{\{\\
        "Human Version Strength": "xx",\\
        "Suggestion for AI Version": "xx",\\
    \}\},\\
    \{\{\\
        "Human Version Strength": "xx",\\
        "Suggestion for AI Version": "xx",\\
    \}\}\\
\rbrack\\
\texttt{\`}\texttt{\`}\texttt{\`}
\end{tcolorbox}
% \newpage
% \input{checklist}
%%%%%%%%%%%%%%%%%%%%%%%%%%%%%%%%%%%%%%%%%%%%%%%%%%%%%%%%%%%%

\end{document}